\documentclass[10pt,journal,compsoc]{IEEEtran}

\usepackage{amsmath}
\usepackage{times}
\usepackage[utf8]{inputenc}
\usepackage{graphicx,subcaption}
\usepackage{color}
\usepackage[dvipsnames]{xcolor}
\usepackage{multirow}
\usepackage{amsfonts}
\usepackage{amssymb}
\usepackage{mathrsfs}
\usepackage{caption}
\usepackage{booktabs}
\usepackage{url}
\usepackage{mathrsfs}
\usepackage{fourier}

\usepackage{algorithm}
\usepackage{algpseudocode}
\usepackage{tabularx,booktabs}
\newcolumntype{C}{>{\centering\arraybackslash}X} 
\usepackage{lipsum}
\usepackage{etoolbox}
\usepackage{tikz}
\usetikzlibrary{tikzmark}
\usetikzlibrary{calc}
\usepackage{lscape}

\usepackage{rotating}
\usepackage{bbm}
\usepackage{latexsym}
\newtheorem{definition}{Definition}

\usepackage{mathtools} 
\usepackage{booktabs} 
\usepackage{tikz} 

\ifCLASSOPTIONcompsoc
  \usepackage[nocompress]{cite}
\else
  \usepackage{cite}
\fi

\begin{document}
\title{Quantifying Model Uncertainty for Semantic Segmentation using Operators in the RKHS}

\author{Rishabh~Singh,~\IEEEmembership{Student Member,~IEEE,}
        and~Jose~C.~Principe,~\IEEEmembership{Life~Fellow,~IEEE}
\IEEEcompsocitemizethanks{\IEEEcompsocthanksitem R. Singh is with the Department
of Electrical and Computer Engineering, University of Florida, Gainesville,
FL, 32607.\protect\\
E-mail: rish283@ufl.edu
\IEEEcompsocthanksitem J. Principe is with the Department
of Electrical and Computer Engineering, University of Florida. E-mail: principe@cnel.ufl.edu}}

\IEEEtitleabstractindextext{
\begin{abstract}
Deep learning models for semantic segmentation are prone to poor performance in real-world applications due to the highly challenging nature of the task. Model uncertainty quantification (UQ) is one way to address this issue of lack of model trustworthiness by enabling the practitioner to know how much to trust a segmentation output. Current UQ methods in this application domain are mainly restricted to Bayesian based methods which are computationally expensive and are only able to extract central moments of uncertainty thereby limiting the quality of their uncertainty estimates. We present a simple framework for high-resolution predictive uncertainty quantification of semantic segmentation models that leverages a multi-moment functional definition of uncertainty associated with the model's feature space in the reproducing kernel Hilbert space (RKHS). The multiple uncertainty functionals extracted from this framework are defined by the local density dynamics of the model's feature space and hence automatically align themselves at the tail-regions of the intrinsic probability density function of the feature space (where uncertainty is the highest) in such a way that the successively higher order moments quantify the more uncertain regions. This leads to a significantly more accurate view of model uncertainty than conventional Bayesian methods. Moreover, the extraction of such moments is done in a \textit{single-shot} computation making it much faster than Bayesian and ensemble approaches (that involve a high number of forward stochastic passes of the model to quantify its uncertainty). We demonstrate these advantages through experimental evaluations of our framework implemented over four different state-of-the-art model architectures that are trained and evaluated on two benchmark road-scene segmentation datasets (Camvid and Cityscapes).
\end{abstract}
}

\maketitle


\noindent Semantic segmentation refers to the task of assigning each pixel of an image with a predefined category label. This has widespread applications in diverse domains such as autonomous driving, computer aided medical diagnosis, human-machine interaction systems, computational photography, image search interfaces, etc. Compared to the traditional computer vision tasks of classification (where the goal is to assign a single label to an entire input image) and object detection (where the goal is to detect locations of objects in an image using bounding boxes), semantic segmentation is a far more challenging task because it requires object location, categorization \textit{and} delineation of their boundaries from each other. Therefore, it requires an efficient utilization of both low-level features and high level semantics associated with images.\par

The advent of deep learning algorithms have made them the predominant choice of algorithms for pattern recognition in many applications of computer vision, including semantic segmentation. Particularly, the introduction of fully convolutional networks (FCN) by \cite{fcn} mainstreamed the use of CNN based architectures, as hierarchical feature learning visual models, for semantic segmentation. Subsequently, numerous more feature enhancing architectures were introduced that significantly improved upon this model architecture such as U-Net \cite{unet}, RefineNet \cite{refinenet}, DeepLabV3 \cite{dlv3} and region based CNN called ReSeg \cite{reseg}. To mitigate the lack of global context information in traditional FCNs, many methods were introduced such as Conditional Random Fields or CRFs \cite{crf}, DilatedNet \cite{dilnet}, which involves different scales of convolutions, and Global Context Embedding Network (GCENet) \cite{gce}.



Despite the progress of deep learning methodologies, semantic segmentation remains a very difficult task in computer vision due to the many practical challenges involved. One such challenge is the lack of high quality and well labelled training data available and also the high costs of generating them. For example, it takes at-least 1.5 hours for an expert to fully label \textit{a single} Cityscapes dataset image \cite{s3}. Another major roadblock is the requirement of semantic segmentation models to be not only highly accurate, but also be very fast in their implementations. In general, any implementation of such models should be able to cope with a typical camera rate of 25 frames per second (FPS). However, current state-of-the-art architectures like FCNs have a rate of 10 FPS for processing even very low resolution images like those found in PASCAL VOC dataset. Other architectures like CRFasRNN have rates as slow as 2 FPS \cite{s1, s2}. Other challenges include susceptibility of models to adverse testing conditions (bad lighting, reflections, fog, rain etc.), occlusions, distributional shifts in data and domain gaps between different datasets \cite{s3}. Moreover, there are also more fundamental problems faced by deep learning algorithms for the purpose of semantic segmentation such as lack of context knowledge during training and overfitting. These problems necessitate quantification of performance trustworthiness in deep learning models for semantic segmentation, especially in safety critical applications like autonomous vision (used in self-driving vehicles), medical diagnosis, etc. Accurate and fast uncertainty quantification methods can thus be very useful in this regard since they allow the practitioner to know how much to trust the results of such models.\par

Most of the advances in uncertainty quantification methods in machine learning have been limited to traditional Bayesian and ensemble based techniques \cite{be1,be2,be3,be4,be5,be6,be7,be8,be9,be10,gal,laks1,laks2}. Over the past 2-3 years there have also been many implementations of such methods for quantifying semantic segmentation uncertainty \cite{bunet,dist,mukh,vid,rob}. \cite{mukh} for instance, creates a benchmark of performance for Monte Carlo (MC) dropout and concrete dropout methods of Bayesian inference on DeepLab-v3 model of semantic segmentation on the Cityscapes dataset. \cite{bunet} develops Bayesian U-Net model which utilizes MC dropout inference over the U-Net segmentation model to quantify uncertainty in the segmentation of aerial and satellite images. \cite{rob} also uses MC dropout inference combined with a new weighted cross-entropy loss function to improve performance of segmentation of remote sensing images by deep CNNs. The Bayesian formulation of model uncertainty involving the inference of probability distribution of model weights has therefore been well established as the gold standard for uncertainty estimation. Although mathematically very elegant, such methods have been infamous for being highly iterative and hence computationally expensive to implement in practice, especially when it comes to larger models and datasets. Moreover Bayesian methodologies are only able to quantify the central moments of uncertainty (mean and variance) which vastly limits the precision of their uncertainty estimates. This become especially disadvantageous in the particular application of semantic segmentation where high implementation speed and high accuracy are both very important. \par

More recently and more relevant to our work, an interesting line of research work attempts to overcome the computational disadvantages of traditional Bayesian and ensemble methods by estimating epistemic uncertainty of models in a \textit{single-pass} and by treating the weights deterministically \cite{dum1,dum2,dum3,dum4,ddu1,ddu2}. In particular, \cite{ddu1} introduced Deep Deterministic Uncertainty (DDU) method which showed that traditionally trained deterministic neural networks enforced with smoothness and sensitivity on the feature-space capture epistemic uncertainty intrinsically through the feature space densities. \cite{ddu2} further extended this for semantic segmentation uncertainty quantification as well. Specifically, authors show that distance (or density) between unseen test-data and training data in the feature space serves as reliable proxy for epistemic uncertainty thereby overcoming the drawback of softmax entropy (which only captures aleatoric uncertainty). However \cite{dumbad} very recently showed that all of these methods, typically referred to as deterministic uncertainty methods (DUM), fail to capture uncertainty estimates as reliably as typical Bayesian approaches in situations involving distributional shifts and out-of-distribution data as well as in semantic segmentation tasks.\par

\section{Contributions}
In this paper, we propose a method for high-resolution quantification of model uncertainty through a single-shot and multi-moment decomposition of the probability density function (PDF) of the training data's representation in the model's feature space (specifically the pre-activation layer). We propose to do this by using a recently introduced framework for uncertainty decomposition of data called the Quantum Information Potential Field (QIPF) \cite{me1,me2, me3}, which leverages physics inspired gradient-flow operators over the projection of data in a reproducing kernel Hilbert space (RKHS) to decompose its PDF in terms on multiple and systematically located orthogonal functionals that signify the tail-regions of data PDF where the uncertainty is highest. We aim to extract similar uncertainty functionals of the training data in the feature space of the trained model so that the evaluation of an unseen (test) sample's feature space representation in these uncertainty functionals would give a precise estimate of the model uncertainty associated with that sample (based on the index of the uncertainty functional/moment which has the highest value associated with the sample.) Our approach (depicted in Fig. 1) therefore falls in the category of deterministic uncertainty methods. We summarize its advantages as follows:

\begin{figure}[!t]
    \centering\includegraphics[scale = 0.15]{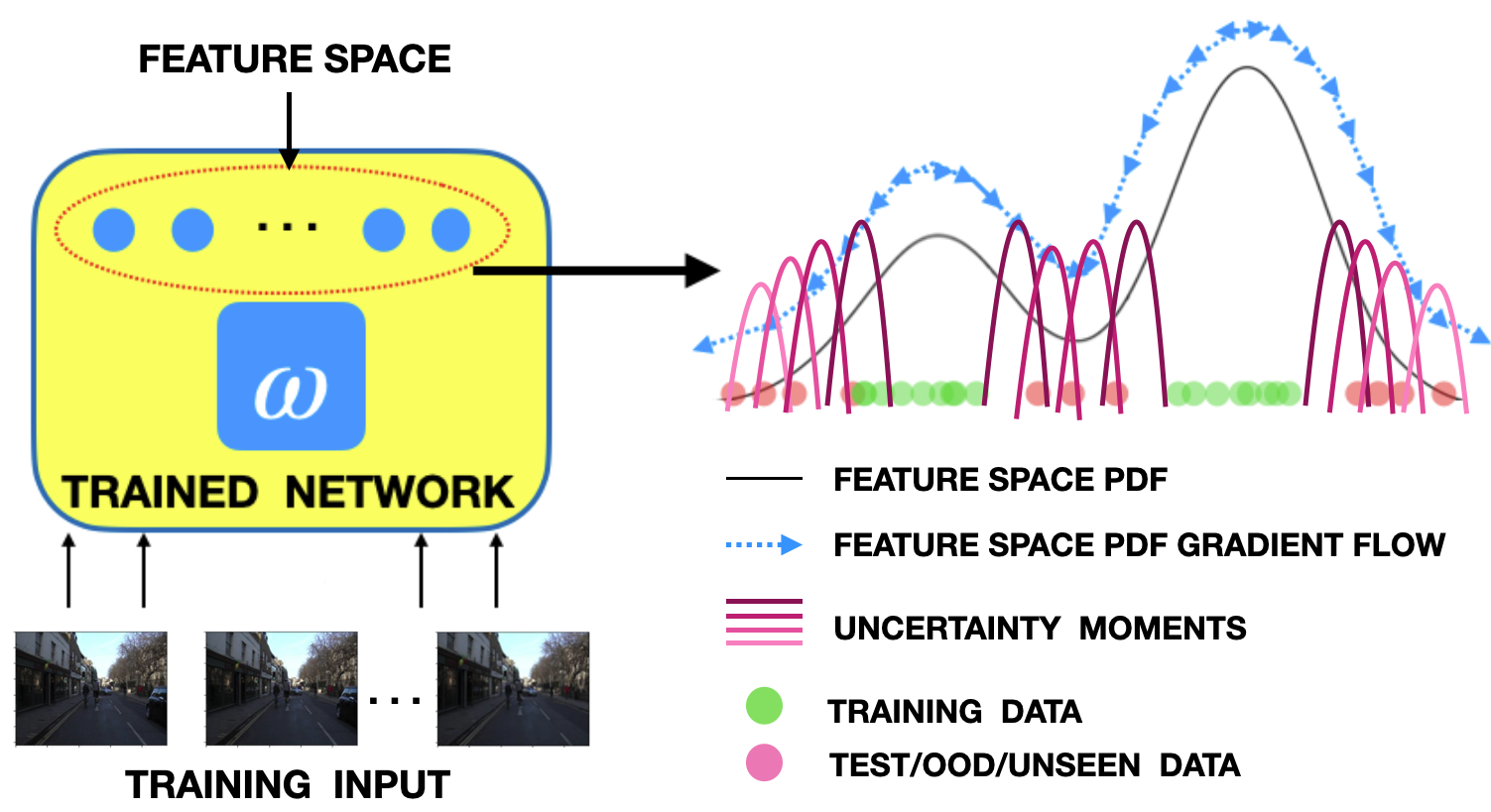}
    \caption{Overview of QIPF Framework for semantic segmentation: The feature space PDF is first quantified in the RKHS. Uncertainty moments are obtained through a decomposition of the gradient flow of feature space PDF. They quantify regions of the PDF with low sample density where out-of-distribution (OOD) samples are likely to occur.}
    \label{fm}
\end{figure}

\begin{itemize}
    \item \textbf{Speed \& Precision:} The QIPF framework has numerous advantages over traditional Bayesian techniques:
    \begin{itemize}
        \item The functional definition of data PDF in the RKHS enables a principled decomposition of the PDF in terms of multiple local uncertainty moments (through the use of physics inspired formulations). This enables a significantly more precise quantification of uncertainty than Bayesian methods (that are only limited to central moments of uncertainty).
        \item  Unlike Bayesian approaches which require hundreds of stochastic forward passes, QIPF enables a single-shot computation of uncertainty, making it much faster and enabling it to scale better to larger models and datasets.
    \end{itemize}
    These advantages of speed and accuracy are particularly important in the application of semantic segmentation (perhaps more than in other computer vision tasks).
    \item \textbf{Data Efficiency:} The QIPF leverages the Gaussian RKHS which has many elegant mathematical properties making it a universal injective function for estimating the data PDF while also being highly data efficient (ability to infer PDF using limited number of samples). This makes our approach particularly valuable since well-labelled data is usually limited in this application domain.
    \item \textbf{Generalization:} We present a method that is non-intrusive to model training and hence can be generalized to all models used for semantic segmentation.
\end{itemize}

Other contributions of this paper are as follows:
\begin{itemize}
\item We provide a systematic description of the QIPF framework from the perspective of functional analysis in the RKHS and use perturbation theory to derive the functional operator for uncertainty quantification (UQ). The paper provides a new intuition into how a \textit{potential field} viewpoint of the model feature space can be used to describe the degree of uncertainty of model in the local neighborhood associated with the output in terms of multiple moments.
\item We analyze the performance of the QIPF against baselines UQ approaches using four different state-of-the-art segmentation models (FCN, PSPNet, SegNet, and U-Net) trained on two popular scene segmentation datasets (camvid and cityscapes). We use several well-known metrics for our analysis.
\end{itemize}

\section{Problem Formulation \&  Background}
We describe the problem setup as follows. Consider a neural network model trained on a dataset $\mathbf{D} = \{\mathbf{x}_{i}, y_{i}\}_{i=1}^N$ consisting of $N$ i.i.d input-output training pairs $\{\mathbf{x},y\}$ where $\mathbf{x}_i \in \mathbb{R}^d$ represents $d$-dimensional input features and $y_i \in \{1...k\}$ represents the target labels where $i$ represents the sample number. A neural network model can be treated as a model that learns the conditional distribution $p(y|\mathbf{x}, \mathbf{w})$. Particularly if the neural network is assumed deterministic, the problem is then that of finding an optimal set of weights $\mathbf{w}$ that correspond to the optimal solution of a maximum log-likelihood estimation problem given by $p(\mathbf{D}|\mathbf{w}) = \Pi_ip(y_i|\mathbf{x}_i, \mathbf{w})$. The Bayesian approach treats the model weights probabilistically and hence attempts to solve the following maximum a-posteriori estimation problem:

\begin{equation}
    p(\mathbf{w}|\mathbf{D}) = \frac{p(\mathbf{D}|\mathbf{w})p(\mathbf{w})}{\int{p(\mathbf{D}|\mathbf{w})p(\mathbf{w})d\mathbf{w}}}
\end{equation}

This inference problem is then followed my prediction phase involving estimation of the posterior predictive PDF of the model given by:

 \begin{equation}
     p(y^*|\mathbf{x}^*, \mathbf{D}) = \int{p(y^*|\mathbf{x}^*, \mathbf{w})p(\mathbf{w}|\mathbf{D})d\mathbf{w}}
 \end{equation}

 Such an inference problem is intractable to solve optimally in modern applications because of the difficulty in computing the marginal likelihood. Further, the posterior is also computationally expensive to estimate since it requires marginalization over weights. Hence most Bayesian methods involve ways to approximate solutions to (1) and (2) using stochastic iterative approaches intrinsically involving multiple realizations of the model.\par
 
 On the other hand, the Deep Deterministic Uncertainty (DDU) method introduced by \cite{ddu1} involves computing the feature space means and covariance matrix associated per pixel or class which are then used to fit a Gaussian Discriminant Analysis (GDA). More specifically, let $z$ be the feature representation of training data $\mathbf{x}$ so that $z = f_\mathbf{w}(\mathbf{x})$. DDU involves computing the feature density $p(z)$ by marginalizing over all pixels or classes denoted by $c$ as follows:

 \begin{equation}
     p(z) = \sum_c{p(z|c)p(c)}
 \end{equation}

 where $p(z|c)$ obtained from GDA ($p(c)$ being computed from training samples) quantifies the model uncertainty associated with pixel/class $c$.\par

 Unlike above approaches, our goal is to express $p(z)$ directly in terms of high-resolution uncertainty functionals/moments. Toward this end, we start with the projection of $z$ into the Gaussian RKHS using a metric called the information potential field (IPF) to quantify its intrinsic PDF as a functional. This is denoted by $\psi_z(.)$ and is a first initial approximation/proxy to $p(z)$ so that $\psi_z(z^*) \sim p(z|c)$ where $z^*$ is the feature space representation of samples of a class $c$. Using the QIPF framework, we aim to intrinsically quantify uncertainty associated with $\psi_z(.) \sim p(z)$ by evaluating its variability in the local neighborhood of the evaluation point (test-sample) $z^*$. We denote this local variability/dynamics of $p(z)$ around $z^*$ as $\psi_z(z^* + \nabla{z^*})$. The intuition here is that a lack of training samples corresponding to the local feature space around a test-input sample will lead to high variability in the predictions within that space (because of the sparseness of training samples in that space). In order to achieve this, we use gradient based operators inspired from quantum physics to decompose $\psi_z(z^* + \nabla{z^*})$ in terms of uncertainty moments $H^i(\psi_z(.))$ where $i \in \{1..m\}$ denotes the moment index so that successively higher order uncertainty moments $H^0(\psi_z(.)), H^1(\psi_z(.))...H^m(\psi_z(.))$ quantify the more uncertain regions of $\psi_z(.)$. In other words, we cluster $p(z)$ systematically in terms of uncertainty.\par

 \begin{figure*}[!t]
    \centering\includegraphics[scale = 0.29]{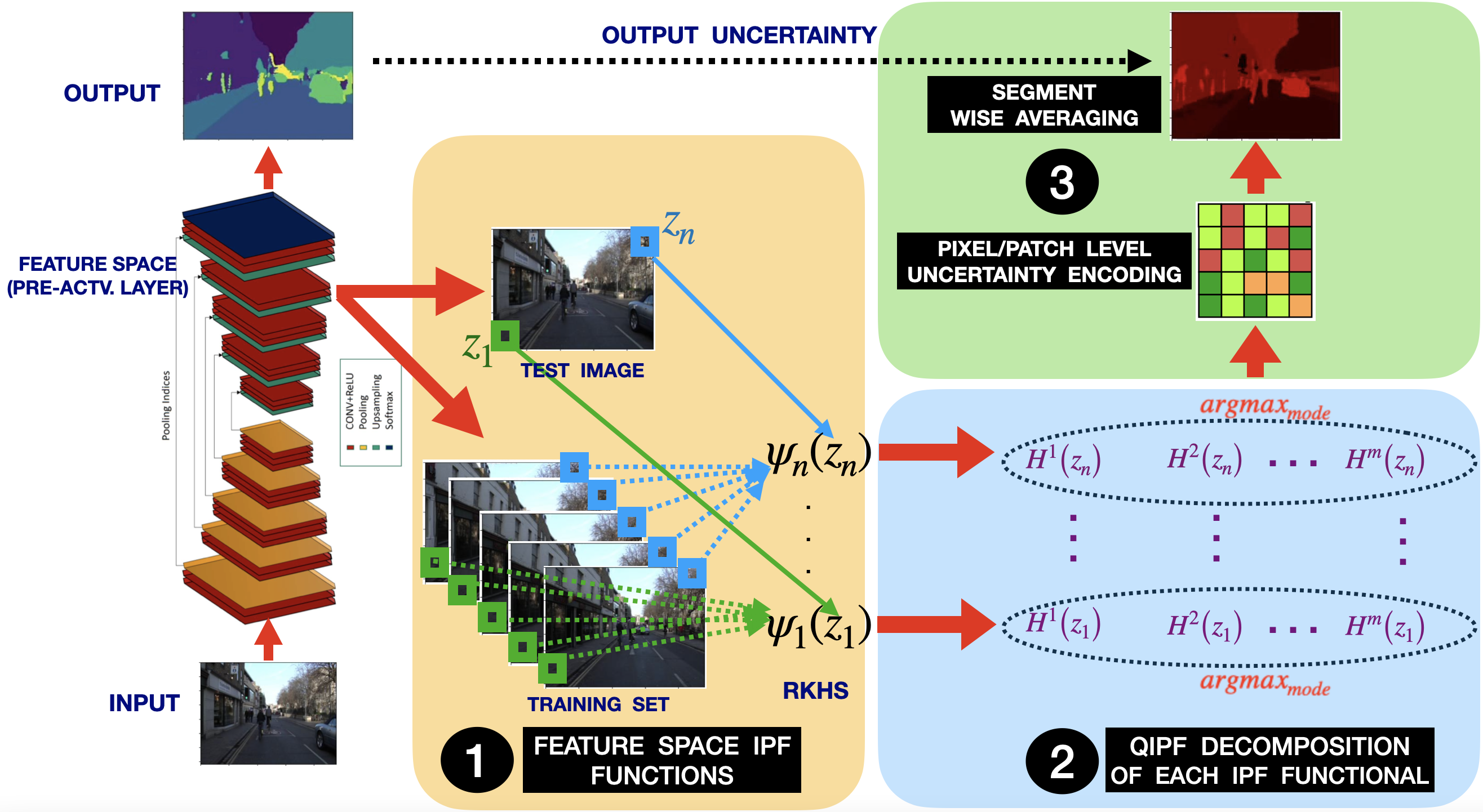}
    \caption{QIPF implementation for semantic segmentation uncertainty quantification: 1. For each pixel/patch, feature space IPF functional is constructed using the training set. Each pixel/patch of the test frame is evaluated on the corresponding IPF. 2. QIPF decomposition of each IPF functional. 3. Pixel/patch level uncertainty encoding obtained for the test frame by taking the index of the moment with highest value for each pixel/patch. Segment -wise averaging to obtain the unceratinty map.}
    \label{fm1}
\end{figure*}

\section{QIPF Uncertainty Framework}
 We begin the description of the QIPF framework for semantic segmentation uncertainty quantification by first formally defining the information potential field (IPF) associated with the feature space representation of training data as follows:

\begin{definition}[Feature Space Information Potential Field]
Given a non-empty set $X = \Omega \subset \mathbb{R}^d$ representing a random variable associated with a pixel/patch of the input training image data, its representation in the model's feature space given by $Z = f_\mathbf{w}(X)$ and a symmetric non-negative definite kernel function $K: Z \times Z \rightarrow \mathbb{R}$, the information potential field (IPF) is an estimator of the PDF of $Z$ in the RKHS from $n$ samples ($z_1, z_2, ..., z_n$) evaluated at any test-point $z^* \in Z$ as

\begin{equation}
 \psi_z(z^*) = \frac{1}{n}\sum_{t=1}^{n}K(z_t, z^*).
 \label{ipf}
\end{equation}

\end{definition}

The IPF is the RKHS equivalent of the Parzen's window method \cite{parz} and is therefore a non-parametric estimator of a continuous density function $f_Z(z^*)$ in an asymptotically unbiased form directly from data (i.e. satisfying the condition that $\lim_{n\to\infty} E[f_Z(\mathbf{z}_n)] = f_Z(\mathbf{z})$, where $n$ is the number of samples). We choose $K$ to be a Gaussian kernel $G$ (with kernel width $\sigma$) without loss of generality towards other symmetric non-negative definite functions. The IPF is therefore a reliable RKHS based unbiased functional estimator of the feature space PDF defined empirically by samples. The potential field analogy of IPF is simple to explain: If the projected Gaussian function centered at a data sample is associated with a unit mass, then it is easy to prove that the interactions of samples projected in RKHS define a potential field resembling a gravitational field specified by the inner product. So, from a physics point of view the IPF, $\psi_z(.)$, can be thought of as a wave-function created by the density of the projected data samples.\par

In this work, we consider the model's feature space as the representation of the input in its pre-activation layer. Much like DDU, there is flexibility in what we consider to be an input sample (i.e. a pixel, patch or class) though we depict the framework considering the input to be in pixel level for clarity in the rest of the paper.\par

Our approach calls for extraction of more specific information i.e. the variability of $\psi_z(.)$ in the local neighborhood of point of evaluation $z^*$, which is unfortunately not readily quantified by Bayesian or probability theory. The functional form of $\psi_z(.)$ in the RKHS allows us to use gradient based operators, particularly the Laplacian operator, $\nabla^2$, to measure the local gradient flow of $\psi_z(.)$ across the space of test-sample $z^*$. Such a quantification of the local behavior of a functional by using operators is most efficiently implemented in quantum mathematics (specifically through the construction of a Schr\"odinger's equation) in terms of multiple moments/eigenfunctions (that provide high-resolution picture of the local dynamics of the functional). Therefore, towards the goal of expressing local gradient flow of $\psi_z(.)$ in a moment-decomposable form, we define a Schr\"odinger's equation given by $H_{z}\psi_{z}(.) = E_z(.)\psi_z(.)$, where $H_z$ is a Hamiltonian operator measuring the local rate of change of $\psi_z(.)$ (feature space IPF) to quantify its uncertainty and $E_z$ denotes associated eigenfunction (or the energy-state). We then solve this Schr\"odinger's equation to obtain a solution of the local variability of $\psi_z(.)$ in terms of multiple systematic local moments by using well-known approach in quantum mathematics called the perturbation theory \cite{pert}. The general perturbation theory is described in appendix A and we provide the solution of the specific Schr\"odinger's equation associated with $\psi_z(.)$ using perturbation theory in appendix B. We also refer the reader to \cite{me3} (particularly section 4 and 5) for more context. In summary, we add a Laplacian based perturbation Hamiltonian $H_p = -\frac{\sigma^2}{2}\nabla_z^2$ (scaled appropriately by the kernel width of the IPF, $\sigma$) to the original Hamiltonian to modify the Schr\"odinger's equation as $H\psi_z(.) = \bigg(H_z(.) - \frac{\sigma^2}{2}\nabla_{z}^2\bigg)\psi_z(.) = E_z(.)\psi_z(.)$ thereby making it closely evaluate the dynamics of $\psi_z(.)$ in the local neighborhood of any evaluation point $z^*$. Rearranging the terms as an expression of $H_z(.)$ and evaluating it on $z^*$, we obtain:


\begin{equation}
    H_z(z^*) = E_z(z^*) + (\sigma^2/2)\frac{\nabla_z^2\psi_z(z^*)}{\psi_z(z^*)}.
\end{equation}

We call this as the quantum information potential field (QIPF) of the feature space (or feature space QIPF). This is therefore a functional operator in the RKHS that evaluates the normalized Laplacian of the feature space IPF, $\psi_z(.)$, to describe its local dynamics (rate of change). It has a small value in parts of the RKHS that have high density of samples and increases its value in parts of the space that are sparse in projected feature space data.\par

Following perturbation theory further, one can decompose the feature space-QIPF, $H_z(.)$, in terms of multiple uncertainty moments via projections of feature space IPF, $\psi_z(.)$, along \textit{successively higher order Hermite polynomial spaces}. We summarize this QIPF decomposition formally as follows and refer the reader to appendix B for its derivation details.

\begin{definition}[Feature Space QIPF Decomposition]
Consider feature space information potential field of a trained model to be given by $\psi_z(.) = \frac{1}{n}\sum_{i=1}^{n}G_\sigma(z_i, .)$, where $z_i$ refers to feature space representation of the $i^{th}$ training sample and $G$ denotes Gaussian kernel with its kernel width $\sigma$. The model's uncertainty associated with a particular test-input sample $\mathbf{x}^*$, with a corresponding feature space projection as $z^*$, can be described by an ordered set of $m$ decomposed feature space-QIPF moments $\{H_z^1(z^*), H_z^2(z^*), ..., H_z^m(z^*)\}$, the $k^{th}$ moment of which is given by the expression:

\begin{equation}
    H_z^k(z^*) = E_z^k(z^*) + (\sigma^2/2)\frac{\nabla_z^2\psi_z^k(z^*)}{\psi_z^k(z^*)}
    \label{qipf}
\end{equation}
where, \\
$\psi_z^k(.)$: $k^{th}$ order Hermite function (normalized) projection of the feature space IPF $\psi_z(.)$. \\
$\nabla_z^2$: Laplacian operator acting with respect to model feature space $z$.\\
$E_z^k(.) = -min(\sigma^2/2)\frac{\nabla_z^2\psi_z^k(.)}{\psi_z^k(.)}$: lower bound to ensure each moment is positive.
\end{definition}

$H_z^0(z^*)$, $H_z^1(z^*)$, $H_z^2(z^*)$ ... are thus the successive local moments of model uncertainty (called QIPF moments) that induce anisotropy in the model's feature PDF space and represent functional measurements at $z^*$ corresponding to \textit{different degrees of heterogeneity of feature space PDF} at $z^*$. Definitions 1-2 therefore fully summarize the extension of QIPF uncertainty decomposition framework for semantic segmentation models. 

\section{Implementation Method}
The overall implementation of the QIPF framework is depicted in Fig. 2. We start by considering a trained semantic segmentation model with access to its feature space, specifically the final pre-activation (or pre-softmax) layer output. The implementation steps are described as follows:
\begin{itemize}
    \item We first construct a feature space IPF functional (\ref{ipf}) for each pixel location by using the feature space representation of all corresponding pixels in the training set. The feature space value of the test image pixel is then evaluated using the corresponding IPF functional.
    \item We then perform QIPF decomposition (\ref{qipf}) to extract $m$ uncertainty moments/functionals for each IPF functional.
    \item Uncertainty for each test image pixel is then defined as the index of the uncertainty moment (from the set of $m$ moments corresponding to that pixel) having the largest value (i.e. argmax$_k(H^k(z_n))$ for the $n^{th}$ pixel where $k$ denotes the moment index).
    \item Segment-wise averaging can then be performed on the obtained pixel-wise uncertainty map.
\end{itemize}
We remark here that the framework can also be implemented on patches/classes instead of individual pixels.

 \begin{figure}[!t]
    \centering\includegraphics[scale = 0.18]{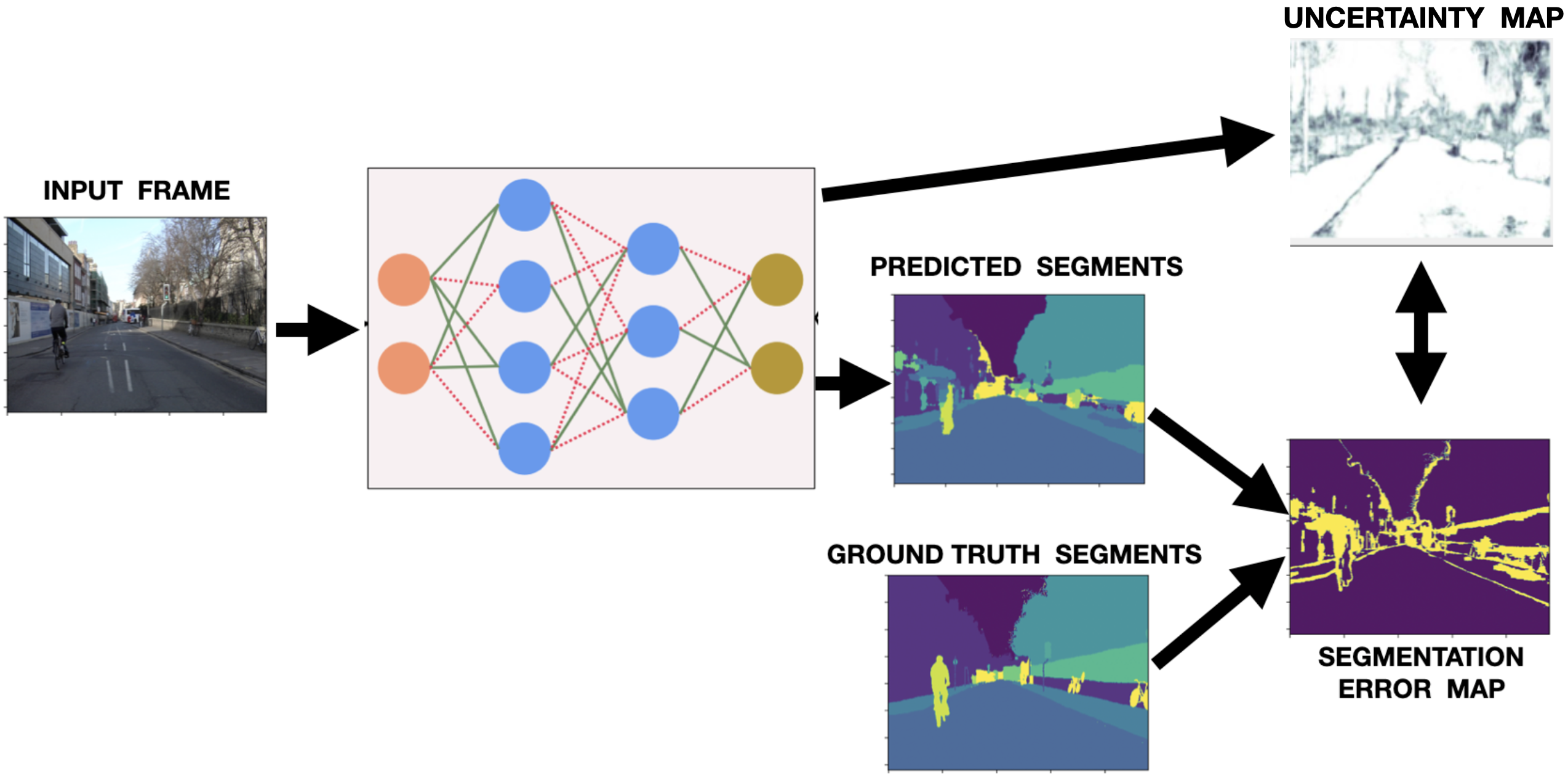}
    \caption{Evaluation method: Uncertainty map is compared with the segmentation error map.}
    \label{fm2}
\end{figure}

\section{Evaluation Methods \& Metrics}
Our method for evaluating and comparing the quality of uncertainty estimates of different methods is depicted in Fig. 3. A test input frame is fed to a trained model which outputs its segmentation map. This predicted segmentation map is then compared with the ground-truth segment map corresponding to the input in order to obtain a segmentation error map. The uncertainty map corresponding to the input is finally evaluated by measuring how much it corresponds with the segmentation error map.\par

An ideal uncertainty map should show high confidence for for patches/pixels that are accurately segmented and low confidence for patches/pixels that are inaccurately segmented. In order to assess the uncertainty estimated based on this principle, we use three metrics proposed by \cite{mukh}  that rely on a confusion matrix containing the number of patches that are accurate and certain ($n_{ac}$), accurate and uncertain ($n_{au}$), inaccurate and certain ($n_{ic}$) and inaccurate and uncertain ($n_{iu}$). These metrics are described as follows:

\begin{enumerate}

    \item \textbf{Patch Accuracy (PA):} Probability that the model is accurate in its segmentation of the patch given that it is confident:
    \begin{center}
        p(\textbf{accurate | certain})  $ = \frac{n_{ac}}{n_{ac} + n_{ic}}$
    \end{center}

    \item \textbf{Patch Uncertainty (PU):} Probability that a model is uncertain about the segmentation of the patch given that the segmentation is inaccurate:
    \begin{center}
        p(\textbf{uncertain | inaccurate})  $ = \frac{n_{iu}}{n_{ic} + n_{iu}}$
    \end{center}

    \item \textbf{PAvsPU:} Combination of both cases of (accurate, certain) and (uncertain, inaccurate):
    \begin{center}
        PAvsPU  $ = \frac{n_{ac} + n_{iu}}{n_{ac} + n_{au} + n_{ic} + n_{iu}}$
    \end{center}    
\end{enumerate}

The values of $n_{ac}$, $n_{au}$, $n_{ic}$ and $n_{iu}$ are obtained after setting the uncertainty threshold as $u_{th} = u_{min} + t(u_{max} - u_{min})$, where $t$ is varied in the range $[0, 1]$. $u_{min}$ is the minimum uncertainty value obtained over the training/validation set and $u_{min}$ is the maximum uncertainty value. An ideal uncertainty quantification method should give out uncertainty estimates having high values of all three metrics at all threshold values.

 \begin{figure*}[!t]
    \centering\includegraphics[scale = 0.35]{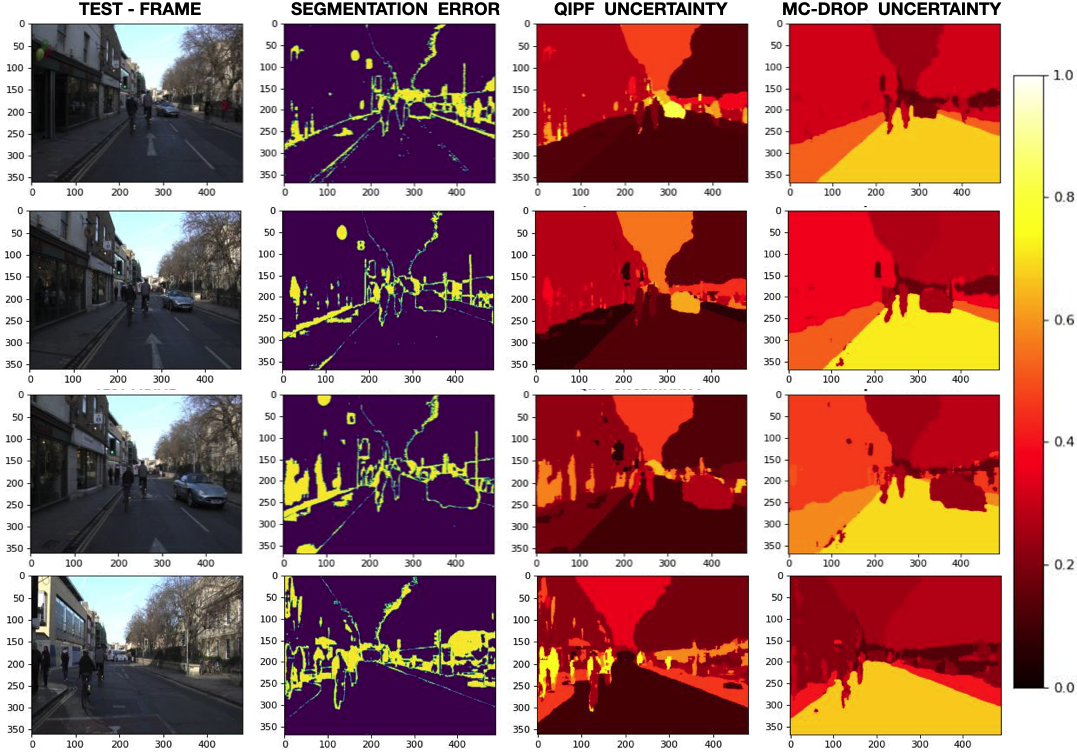}
    \caption{Comparison of QIPF and MC dropout uncertainties with the segmentation error map of a trained FCN-8 model for different test-set frames. QIPF can be seen to produce sharper uncertainty maps that are more correlated with the error map.}
    \label{img1}
\end{figure*}

 \begin{figure*}[!t]
    \centering{
    \includegraphics[scale = 0.3]{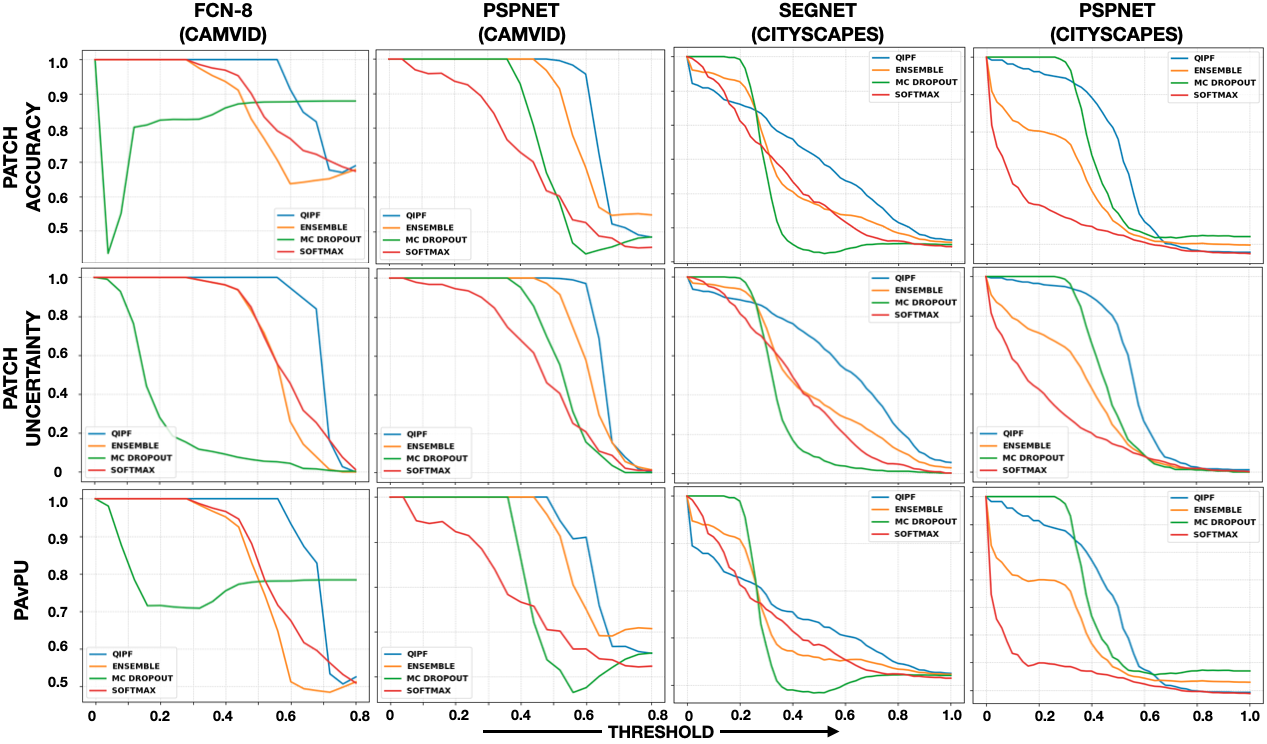}
    \caption{Performance of uncertainty quantification methods implemented on FCN-8 (Camvid), PSPNet (Camvid), SegNet (Cityscapes), PSPNet (Cityscapes) models (from left to right columns respectively). Top row: Patch accuracy vs threshold, Middle row: Patch uncertainty vs threshold, bottom row: PAvsPU vs threshold.}
    \label{img2}
    }
\end{figure*}

\section{Experiments}
Python 3.7 with TensorFlow library is employed to perform all simulations. We compared the QIPF framework three different baseline uncertainty quantification methods namely MC (Monte Carlo) dropout \cite{gal}, the ensemble technique \cite{laks2} and Softmax uncertainty \cite{pear}. We trained four different segmentation models namely FCN-8 \cite{fcn}, SegNet \cite{seg}, PSPNet \cite{psp} and UNet \cite{unet} on two benchmark road scene semantic segmentation datasets: Camvid \cite{cam} and Cityscapes \cite{city}.\par

We implemented the QIPF framework on the trained segmentation networks by following the same procedure as described in section 3 (and Fig. 2), i.e. by constructing IPF functionals for each patch/pixel location using the feature-space representations of the training set, and then evaluating the test-frame patches/pixels on the corresponding IPF functionals followed by QIPF decomposition of each functional into uncertainty modes and segment-wise averaging. We chose the kernel width, $\sigma$, in each network to be a factor of the Silverman bandwidth \cite{sil} associated with the IPF functional of each pixel/patch. The best factor was determined through cross-validation for each network from a held-out batch of the training set. The best factors were found to be 30, 100, 1, and 50 for FCN, PSPNet, SegNet and U-Net models respectively. We extracted 12 QIPF modes of each pixel/patch IPF for all networks.\par 

\begin{table*}[!t]
\centering{
\begin{tabular}{llccc|ccc}
                        &            & \multicolumn{3}{c|}{CAMVID}                                                                                                                           & \multicolumn{3}{c}{CITYSCAPES}                                                                                                                       \\ \hline
                        &            & \begin{tabular}[c]{@{}c@{}}PATCH \\ ACCURACY\end{tabular} & \begin{tabular}[c]{@{}c@{}}PATCH \\ UNCERTAINTY\end{tabular} & \multicolumn{1}{l|}{PAvPU} & \begin{tabular}[c]{@{}c@{}}PATCH \\ ACCURACY\end{tabular} & \begin{tabular}[c]{@{}c@{}}PATCH \\ UNCERTAINTY\end{tabular} & \multicolumn{1}{l}{PAvPU} \\ \hline
\multirow{4}{*}{FCN-8}  & QIPF       & \textbf{0.93}                                             & \textbf{0.85}                                                & \textbf{0.92}              & \textbf{0.89}                                             & \textbf{0.76}                                                & \textbf{0.85}             \\
                        & ENSEMBLE   & 0.86                                                      & 0.69                                                         & 0.81                       & 0.71                                                      & 0.55                                                         & 0.68                      \\
                        & MC DROPOUT & 0.83                                                      & 0.26                                                         & 0.79                       & 0.68                                                      & 0.28                                                         & 0.64                      \\
                        & SOFTMAX    & 0.89                                                      & 0.73                                                         & 0.85                       & 0.66                                                      & 0.31                                                         & 0.56                      \\ \hline
\multirow{4}{*}{SEGNET} & QIPF       & \textbf{0.78}                                             & \textbf{0.66}                                                & \textbf{0.77}              & \textbf{0.74}                                             & \textbf{0.58}                                                & \textbf{0.69}             \\
                        & ENSEMBLE   & 0.66                                                      & 0.65                                                         & 0.69                       & 0.72                                                      & 0.46                                                         & 0.65                      \\
                        & MC DROPOUT & 0.68                                                      & 0.19                                                         & 0.55                       & 0.71                                                      & 0.35                                                         & 0.61                      \\
                        & SOFTMAX    & 0.62                                                      & 0.24                                                         & 0.51                       & 0.72                                                      & 0.41                                                         & 0.63                      \\ \hline
\multirow{4}{*}{PSPNET} & QIPF       & \textbf{0.92}                                             & \textbf{0.83}                                                & \textbf{0.89}              & \textbf{0.81}                                             & \textbf{0.54}                                                & \textbf{0.73}             \\
                        & ENSEMBLE   & 0.90                                                      & 0.75                                                         & 0.86                       & 0.74                                                      & 0.35                                                         & 0.64                      \\
                        & MC DROPOUT & 0.82                                                      & 0.65                                                         & 0.77                       & 0.81                                                      & 0.46                                                         & 0.72                      \\
                        & SOFTMAX    & 0.79                                                      & 0.58                                                         & 0.73                       & 0.68                                                      & 0.23                                                         & 0.56                      \\ \hline
\multirow{4}{*}{U-NET}  & QIPF       & \textbf{0.96}                                             & \textbf{0.90}                                                & \textbf{0.94}              & 0.62                                                      & 0.60                                                         & 0.67                      \\
                        & ENSEMBLE   & 0.86                                                      & 0.71                                                         & 0.83                       & \textbf{0.71}                                             & \textbf{0.65}                                                & \textbf{0.72}             \\
                        & MC DROPOUT & 0.62                                                      & 0.05                                                         & 0.46                       & 0.67                                                      & 0.43                                                         & 0.60                      \\
                        & SOFTMAX    & 0.79                                                      & 0.49                                                         & 0.71                       & 0.62                                                      & 0.34                                                         & 0.53                      \\ \hline
\end{tabular}
}
\caption{Summary of average values of the three evaluation metrics for each model, UQ method and dataset.}
\label{tt1}
\end{table*}

We extracted pixel/patch uncertainties and followed it up by segment-wise averaging in all of the other uncertainty quantification methods as well. For implementing Monte-Carlo (MC) dropout procedure \cite{gal}, we placed a dropout layer (with a small dropout rate) after the pooling layer of each convolutional block in all networks during testing (the best dropout rates were found to be 0.1 for FCN and SegNet and 0.2 for PSPNet and UNet during cross-validation from held-out batch of training set). We implemented 100 stochastic forward passes to the determine the pixel/patch level uncertainty during testing.\par

In the case of ensemble method of uncertainty quantification, we independently trained 8 different versions of each network and computed the standard deviation in the outputs during testing to determine the pixel-level uncertainty.\par

We show the segment-wise uncertainty maps obtained by implementing QIPF and MC dropout on a trained FCN-8 network for four different test-set frames of Camvid dataset in Fig. \ref{img1}. The first column shows the test-frames, second column shows the error map for each test-frame with yellow regions denoting segments where the network made errors in atleast 50\% of the pixels (per segment) during the segmentation task, third column denotes the segment-wise uncertainty quantified by QIPF framework and the fourth column shows the same obtained by MC dropout method. As can be observed in all 5 test cases, the QIPF framework leads to sharper and precise uncertainty maps that visibly correlate much better with the segmentation error map (which is the desired behavior) as compared to the MC dropout method. More examples are provided in appendix C. We quantified this correlation with segmentation error (for each UQ method on all networks implemented on camvid and cityscapes) using the metrics and procedures described in section 4, i.e. patch accuracy (PA), patch uncertainty (PU) and PAvsPU scores at different thresholds on the uncertainty values extracted from the UQ methods. The results obtained for a pair of models for each dataset are shown in Fig. \ref{img2} (left two columns for camvid and right two for cityscapes). As can be seen, the QIPF method consistently obtains higher scores in all of the metrics when compared to other methods, for all four cases. We summarize the average values of the metrics (for all networks and UQ methods on the two datasets) in Table \ref{tt1}. We see here that the QIPF framework achieves significantly better values compared to other UQ methods for most cases. We discuss the computational cost of QIPF in appendix D and also refer the reader to section 9 in \cite{me3}.

\section{Conclusion}
We show in this paper that by implementing a functional decomposition of a trained model's feature space (represented in an RKHS), one can locally measure the interactions between the model's learned representations and its predictions to yield precise estimates of predictive uncertainty in a computationally efficient manner. This is made possible by a set of physics inspired operators (implemented through the Schr\"odinger's formulation) that quantify local anisotropy (variability) of the feature space, which to the best of our knowledge is beyond the capability of statistical methods. Furthermore, the framework of perturbation theory helps us understand the role of the moment decomposition and allows for a possible classification of different types of uncertainty. We applied this RKHS-operator based approach for uncertainty quantification in an important and challenging application of semantic segmentation and showed promising initial results with popular datasets and models. We plan to extend our analysis to bigger models and datasets and also take into account covariate shift and input data corruptions to study our methodology further.

\section{Acknowledgments}
This work was partially supported by DARPA under grant no. FA9453-18-1-0039 and ONR under grant no. N00014-21-1-2345.

\bibliographystyle{IEEEtran}
\bibliography{aaai2022}

\onecolumn
\appendix
\section*{A. PERTURBATION THEORY}

Perturbation theory is an approach in quantum mathematics to express a complicated system as a perturbed version of a similar but simpler base system whose mathematical solution is well known \cite{qtlan}. Specifically, it involves an approximate characterization of the desired system through a \textit{eigenvalue decomposition} (analogical to Taylor series decomposition) of a \textit{locally} perturbed version of a base system. In doing so, the various physical quantities associated with the complicated system (energy levels and eigenstates) get expressed as \textit{corrections} to the known quantities of the base system. In other words, it is a method to extract more intrinsic dynamical information of a system by observing its behavior upon perturbing it.

Suppose that we want to find approximate solutions of a time-independent system characterized by a probability functional, $\psi(.)$ (referred to as wave-function in quantum physics), its associated Hamiltonian $H$ (which is basically a set of operators describing the dynamics of $\psi(.)$) and the energy state of the system, $E(.)$, given by:

\begin{equation}
    H\psi(.) = E(.)\psi(.)
    \label{des}
\end{equation}

Let us assume that there exists a simpler base system (somewhat approximating the desired system above (\ref{des})) with the Hamiltonian, probability functional and energy state as $H_0$, $\psi^{0}(.)$ and $E^0(.)$ respectively, given by:

\begin{equation}
    H_0\psi^{0}(.) = E^{0}(.)\psi^{0}(.)
    \label{simp}
\end{equation}

and whose solutions (i.e. $\psi^{0}(.)$ and $E^0(.)$) are known exactly. Here the superscript 0 denotes the quantities to be associated with an unperturbed system.

Let us consider $H_p$ to be a Hamiltonian which represents a weak perturbation and $\lambda$ to be a dimensionless parameter describing the scale of perturbation ranging between 0 (no perturbation) and 1 (full perturbation), so that the difference between desired system (\ref{des}), described by $H$, and the base system (\ref{simp}), described by $H_0$, is compensated by perturbing the Hamiltonian of (\ref{simp}) by $\lambda H_p$. This leads to an expression of $H$ given by $H = H_0 + \lambda H_p$. One therefore requires to find approximate solutions of the Schr\"odinger's equation associated with the perturbed version of the base system:

\begin{equation}
   H\psi(.) = (H_0 + \lambda H_p)\psi(.)  = E(.)\psi(.)
   \label{q1}
\end{equation}

If we assume $\lambda$ to be sufficiently weak, perturbation theory states that we can express quantities of the desired system ($E$ and $\psi$) as a power-series expansion of the quantities of the base (unperturbed) system. This leads to the following approximate solutions:

\begin{equation}
\begin{split}
E(.) = E^0(.) + \lambda E^1(.) + \lambda^2 E^2(.) + ...  \\
\psi(.) = \psi^0(.) + \lambda\psi^{1}(.) + \lambda^2\psi^{2}(.) + ...
\end{split}
\label{exp}
\end{equation}

where $E^k(.)$ and $\psi^k(.)$ for $k = 0,1,2...$ are the successively higher order orthogonal modes/corrections associated with energy state and probability functional (wave-function) respectively of the base system. This leads to the following expression of the desired system:

\begin{equation}
   H\psi(.) = \bigg(H_0 + \lambda H_P\bigg)\bigg(\psi^0(.) + \psi^1(.) + \psi^2(.) + ...\bigg) = \bigg[E^0(.)\psi^0(.)\bigg] + \bigg[E^1(.)\lambda\psi^1(.)\bigg] + \bigg[E^2(.)\lambda^2\psi^2(.)\bigg] + ...
   \label{q2}
 \end{equation}

\section*{B. SEGMENTATION UNCERTAINTY: DERIVING QIPF MODE DECOMPOSITION}
We work with the feature space PDF of the model where the probability functional is represented by $\psi_z(z^*)$, i.e. the information potential field (Definition 1 in Section 4 of main paper) created by $z$ (feature-space projection of training samples) and evaluated at $z^*$ (feature-space projection of test-input). We aim to extract model uncertainty associated with the test-input by measuring the variability of $\psi_z(z^*)$ in the local neighborhood of $z^*$. We find perturbation theory approach from quantum mathematics (section A of appendix) to be useful here since it allows for quantification of local variability of a probability functional at any evaluation point in terms of multiple moments. Therefore to implement perturbation theory on $\psi_z(.)$, let us assume a base system associated with $\psi_z(z^*)$ as the following Schr\"odinger's equation:

\begin{equation}
    H_z\psi_z(z^*) = E_z(z^*)\psi_z(z^*)
    \label{base}
\end{equation}

Here, $H_z$ is a Hamiltonian, i.e. an operator/functional that measures the local variability of the probability functional $\psi_z(.)$, referred to as the wave-function in quantum mathematics. $E_z(.)$ is the functional that represents the energy-state of the system. Following perturbation theory, we introduce a perturbation Hamiltonian $H_p$ (so that the Hamiltonian corresponding to the new/desired system becomes $H = H_z + H_P$) that shifts the evaluation of $\psi_z(z^*)$ from $z^*$ to its local neighboring space $z^* +  \Delta z^*$. We choose this Hamiltonian to be the Laplacian operator (a local gradient based operator) operating in the space of $z$ (denoted by $\nabla^2_z$). The intensity of the perturbation controls the size of neighborhood around $z^*$ where the analysis is being made and hence is directly dependent on the kernel width, $\sigma$, selected in the IPF formulation $\psi_z(z^*)$. Therefore we choose the perturbation Hamiltonian as $H_p = -\frac{\sigma^2}{2}\nabla_{z^*}^2$ so that desired system Hamiltonian is $H = H_z(.) - \frac{\sigma^2}{2}\nabla_{z^*}^2$. Therefore, the Schr\"odinger's equation corresponding to the desired system becomes:

\begin{equation}
    H\psi(.) = \bigg(H_z(.) - \frac{\sigma^2}{2}\nabla_{z^*}^2\bigg)\psi(.) = E(.)\psi(.)
    \label{s}
\end{equation}

As can be observed, this is identical in form to (\ref{q1}) in section A. Rearranging the terms we get

\begin{equation}
    H_z(.) = E(.) + (\sigma^2/2)\frac{\nabla_z^2\psi(.)}{\psi(.)}
    \label{q5}
\end{equation}

which is the QIPF (quantum information potential field) expression as given in (5) in the main paper (under section 4).

Similar to what is done in section A, we leverage the perturbation theory to express the quantities of the desired system ($E$ and $\psi$) as a power-series expansion of the quantities of the base system ($E_z$ and $\psi_z$):

\begin{equation}
\begin{split}
E(.) = E_z(.) + \lambda E^1_z(.) + \lambda^2 E^2_z(.) + ...  \\
\psi(.) = \psi_z(.) + \lambda\psi^1_z(.) + \lambda^2\psi^2_z(.) + ...
\end{split}
\label{exp}
\end{equation}

where $E_z^k(.)$ and $\psi_z^k(.)$ for $k = 1,2...$ are the successively higher order orthogonal modes/corrections associated with energy-state $E_z(.)$ and probability functional (wave-function) $\psi_z(.)$ respectively of the base system. Implementing this expansion (or moment decomposition) in (\ref{q5}) leads to a decomposition of the functional $H_z(.)$ in terms of its modes $H_z^0(.), H_z^1(.), H_z^2(.), ...$, which when evaluated at $z^*$ becomes:

\begin{equation}
    H_z^k(z^*) = E_z^k(z^*) + (\sigma^2/2)\frac{\nabla_z^2\psi_z^k(z^*)}{\psi_z^k(z^*)}
    \label{qpf}
\end{equation}

for $k = 1, 2, ...$. Here, we take $\psi_z^k(.)$ to be the projection of $\psi_z(.)$ in the $k^{th}$ order Hermite polynomial function to obtain an approximation of the higher order moment. Hermitian polynomials are a family of orthogonal polynomials that frequently appear in power-series expansions and solutions of Schr\"odinger equations \cite{herm} (eg: quantum harmonic oscillator systems) and we find them useful here for obtaining approximate estimations of the QIPF modes. Instead of doing a moment expansion of $E_\mathbf{w}(.)$ here, for practical considerations, we simply let it be a lower bound corresponding to each wave-function moment given by $E_z^k(.) = -min(\sigma^2/2)\frac{\nabla_z^2\psi_z^k(.)}{\psi_z^k(.)}$ to enforce the condition that the corresponding $H_0^k(.)$ is always positive. Hence we have derived the QIPF decomposition expression as defined in definition 2 (section 4) of the main paper.

\section*{C. ADDITIONAL RESULT EXAMPLES (CITYSCAPES DATA SAMPLES)}

\subsection{Cityscapes: FCN QIPF Uncertainty}

 \begin{figure*}[!h]
    \centering\includegraphics[scale = 0.42]{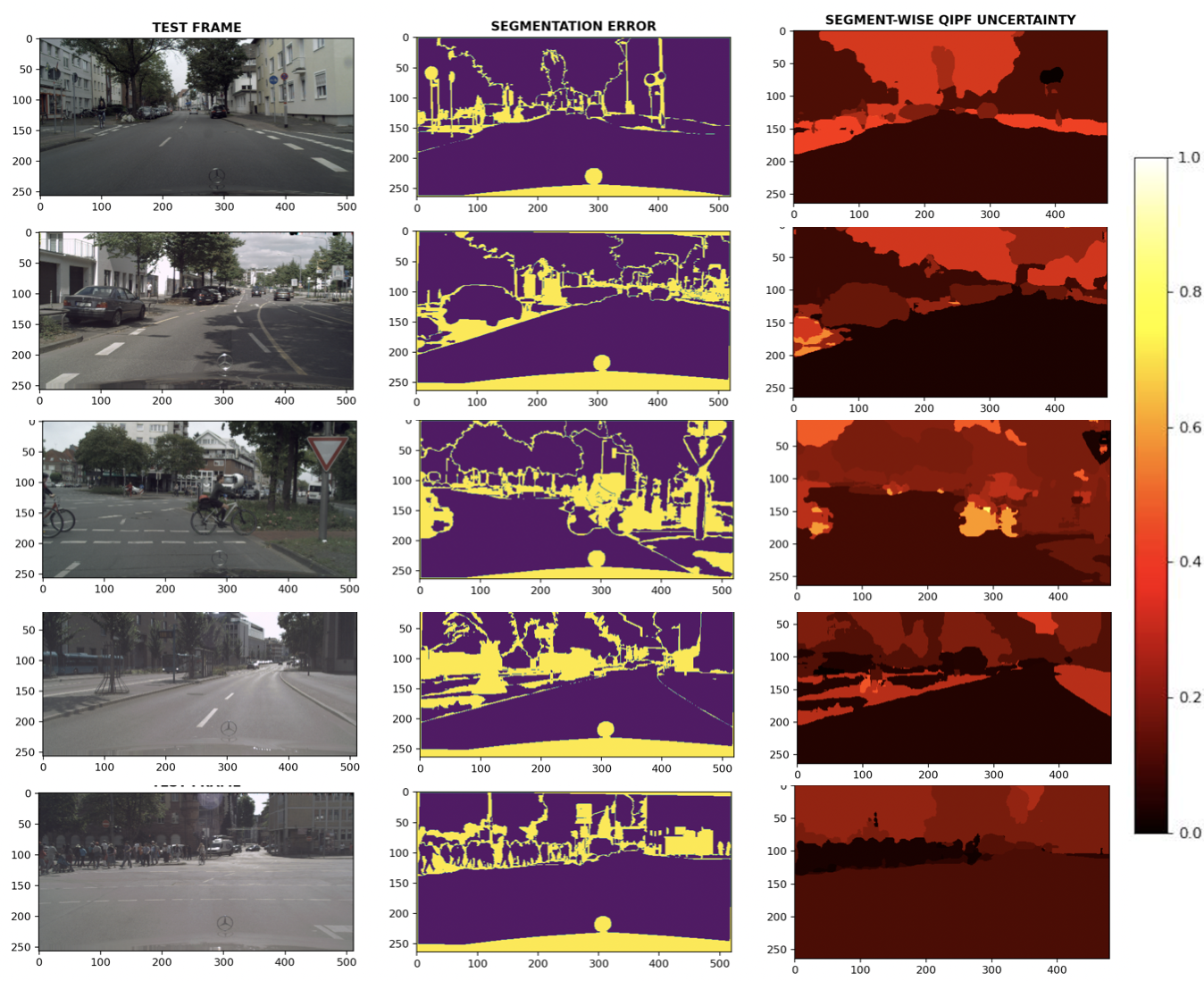}
    \label{imgc1}
\end{figure*}

\subsection{Cityscapes: SegNet QIPF Uncertainty}

 \begin{figure*}[!h]
    \centering\includegraphics[scale = 0.42]{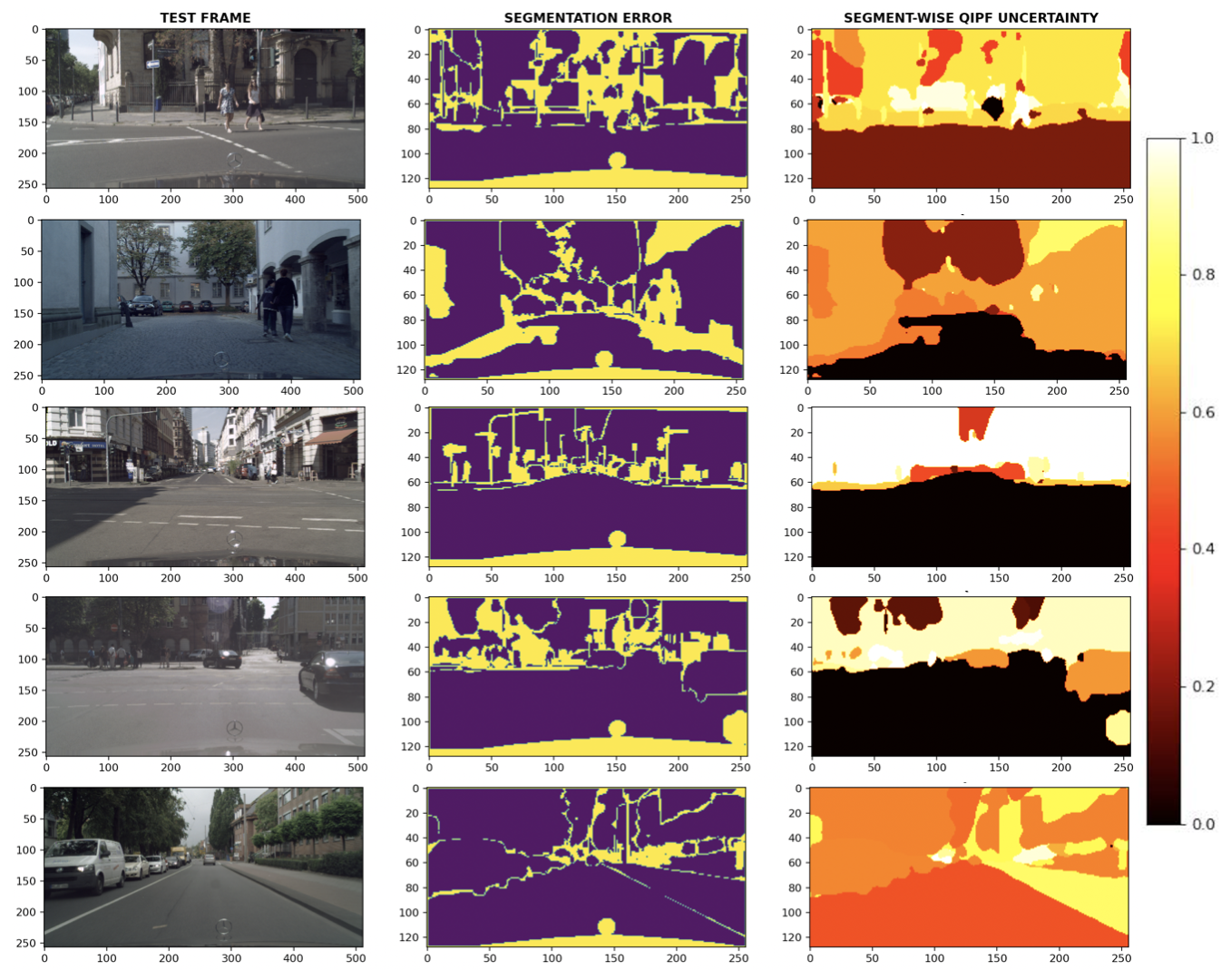}
    \label{imgc2}
\end{figure*}

\section*{D. Computational Cost}
The only computational parts of QIPF are the IPF, $\psi_z(z^*)$, which grows linearly with the size of $z$ and moment extraction growing linearly with the number of moments. The time complexity of the QIPF for each test iteration therefore becomes $\mathcal{O}(n+m)$, where $n$ and $m$ are the number of training inputs and number of moments respectively. In spite of this, QIPF is faster than ensemble or Monte Carlo based methods such as MC dropout which involve very high number of model sampling size for uncertainty estimation. This is because we can considerably downsample the training input to significantly reduce the effective value of $n$, while not compromising much on performance, thanks to the data efficiency of the RKHS (i.e. it is able to quantify PDF even with low number of samples). Furthermore, the value of $m$ is always very low (we used only 12 moments in all experiments) thereby making QIPF much faster.
\vfill

\end{document}